\documentclass[runningheads]{llncs}
\usepackage[T1]{fontenc}
\usepackage{graphicx}
\begin{document}
\title{Can Mental Imagery Improve the Thinking Capabilities of AI Systems?}
%
%
\author{Slimane Larabi}
\authorrunning{S. Larabi}
\institute{USTHB University, BP 32 EL Alia, 16111, Algiers, Algeria\\
\email{s.larabi@usthb.dz}\\
\url{https://perso.usthb.dz/~slarabi/accueil.html} }
\maketitle              
\begin{abstract}
Although existing models can interact with humans and provide satisfactory responses, they lack the ability to act autonomously or engage in independent reasoning. Furthermore, input data in these models is typically provided as explicit queries, even when some sensory data is already acquired.

In addition, AI agents, which are computational entities designed to perform tasks and make decisions autonomously based on their programming, data inputs, and learned knowledge, have shown significant progress. However, they struggle with integrating knowledge across multiple domains, unlike humans.

Mental imagery plays a fundamental role in the brain's thinking process, which involves performing tasks based on internal multisensory data, planned actions, needs, and reasoning capabilities.
In this paper, we investigate how to integrate mental imagery into a machine thinking framework and how this could be beneficial in initiating the thinking process. 
Our proposed machine thinking framework integrates a Cognitive thinking unit supported by three auxiliary units: the Input Data Unit, the Needs Unit, and the Mental Imagery Unit. Within this framework, data is represented as natural language sentences or drawn sketches, serving both informative and decision-making purposes. We conducted validation tests for this framework, and the results are presented and discussed.

\end{abstract}
\section{Introduction}\label{intro}

Creating machines capable of autonomous thinking remains a significant challenge in artificial intelligence (AI). Machine thinking holds great promise for revolutionizing human–machine interaction by enabling machines to perform tasks based on internal data, planned actions, needs, and reasoning processes. Unlike existing AI models that primarily respond to human queries or follow predefined instructions, machine thinking aims to empower systems to simulate autonomous thought, adapt to dynamic environments, and make independent decisions based on contextual demands.

Despite advances in deep learning, computer vision, and natural language processing, current systems fall short of achieving true machine thinking. Prior research has explored key components of machine cognition, including reasoning models~\cite{Sridharan2023}, sensory data integration~\cite{ZHAO2024}, and imagination-driven tasks~\cite{Qi2019}. However, these efforts often lack a unified framework that bridges internal motivations, sensory perception, and reasoning.

As noted in~\cite{Bonnefon2020}, our aim is not to replicate human thought processes but to take inspiration from lifelong experience with how we think and act. Accordingly, in this paper, we present a framework for machine thinking composed of several key elements: the Needs Unit, which captures the machine’s internal needs (scheduled actions, problems to solve, etc.); the Input Data Unit, which captures sensory data such as perceived scenes, sounds, and tactile information; and the Mental Imagery Unit, which enables the simulation and prediction of potential scenarios and provides insight into how AI systems might utilize generative capabilities for reasoning. The Cognitive Thinking Unit (CTU) integrates these auxiliary units to simulate reasoning, plan actions, and generate contextual responses.

To validate the proposed framework, we present experiments across diverse applications: image captioning, which evaluates the system's ability to describe visual inputs; matching needs and context, which tests the system’s ability to align internal motivations with external data; and sentence generation by the Cognitive Thinking Unit (CTU), which demonstrates the framework's reasoning and decision-making capabilities.

The remainder of this paper is organized as follows: Sections~\ref{sect2} and~\ref{sect3} discuss related work in the domains of machine thinking and mental imagery. Sections~\ref{sect4} and~\ref{sect5} describe the proposed framework, detailing its components, functionalities, and how it performs reasoning through the generation of mental images. Section~\ref{sect6} presents the experimental results, and finally, Section~\ref{sect7} concludes the paper.

\section{Related Works to Machine Thinking}\label{sect2}


Machine thinking—the capacity of systems to infer actions or statements from context—is a multifaceted challenge that intersects with several domains of artificial intelligence, including commonsense reasoning, natural language inference (NLI), multi-modal learning, and contextual reasoning.

Commonsense reasoning is a critical yet underdeveloped area in machine intelligence due to the complexity and implicit nature of everyday knowledge ~\cite{Zhu2021}. Approaches include large pretrained models fine-tuned for commonsense tasks (e.g., GPT, T5 ~\cite{Raffel2020}), and structured knowledge bases such as ConceptNet ~\cite{Speer2012}, COMET ~\cite{Bosselut2019}, and ATOMIC ~\cite{Sap2019}, which support inferential reasoning using semantic relationships and textual inferences.

Natural Language Inference (NLI) aims to identify the relationship between a premise and a hypothesis—entailment, contradiction, or neutrality. Datasets like SNLI and MultiNLI ~\cite{Williams2018} have enabled the training of powerful inference models such as BERT and RoBERTa, improving sentence-level understanding.

Multi-modal learning combines heterogeneous inputs such as text and images to perform tasks like Visual Question Answering and cross-modal retrieval. Notable models include CLIP ~\cite{Radford2021}, Flamingo ~\cite{Alayrac2022}, and VilBERT ~\cite{Tan2019}, which align visual and linguistic data to facilitate integrated reasoning.

Contextual and goal-oriented reasoning systems leverage models like BERT and GPT to perform attention-based reasoning over full input sequences ~\cite{Clark2019}. They are central to applications such as autonomous agents and natural language understanding. Additional approaches include rule-based inference systems and multi-modal architectures for action prediction ~\cite{Sridharan2023}.



Despite these advances, current systems remain constrained: They are domain-dependent and often fail to generalize or handle ambiguity ~\cite{Sanyal2023}. They lack deep commonsense reasoning, causality, and temporal understanding ~\cite{Jiang2019}. They propagate data biases and inconsistencies ~\cite{Mavrogiorgos2024}. They exhibit limited creativity and do not autonomously generate novel questions or ideas ~\cite{Stevenson2022}.

Crucially, there is still no framework that enables machines to engage in autonomous thinking—reasoning initiated without an external query. Existing models remain reactive, and while efforts like reinforcement learning agents show promise, they remain task-specific and do not generalize to broader notions of thinking.

\section{Mental Imagery}\label{sect3}

Mental imagery has long been recognized as a foundational component of human cognition, playing a crucial role in memory, planning, decision-making, spatial navigation, and emotional processing ~\cite{Pearson2015}. It supports core cognitive functions by enabling individuals to simulate future scenarios and reconstruct past experiences ~\cite{Moulton2009} ~\cite{Gilbert2007}. These capabilities are essential not only for adaptive behavior but also for understanding and treating mental health conditions.

Recent neurocognitive studies have revealed that mental imagery and perception share overlapping neural substrates. For instance, Stecher et al. ~\cite{Stecher2024} demonstrated that both imagined and perceived natural scenes activate similar cortical patterns, particularly in the alpha frequency band, suggesting that imagery and late-stage perception engage shared neural codes. Key brain regions involved in these processes include the ventromedial prefrontal cortex, anterior hippocampus, posterior parahippocampal cortex, and retrosplenial cortex ~\cite{Monk2020}, reinforcing the biological grounding of visual imagination.

The functional utility of mental imagery extends beyond human cognition into artificial intelligence. Inspired by human intuitive planning, Li et al. ~\cite{Li2023} proposed Simulated Mental Imagery for Planning (SiMIP), a subsymbolic framework that enables robots to "imagine" sequences of actions through internal simulation. SiMIP integrates CNNs and GANs to generate and evaluate visual outcomes of planned actions, thus emulating the role of mental imagery in adaptive behavior and re-planning.

In educational psychology, Commodari et al. ~\cite{Commodari2024} emphasized the role of mental imagery and visuospatial processing in academic performance and creative problem-solving. Their findings advocate for instructional strategies that leverage visual thinking and mental simulation, especially in abstract domains like mathematics and science, to foster deeper cognitive engagement and innovation.

From an AI reasoning perspective, mental imagery offers a significant computational advantage. As shown by Kunda ~\cite{Kunda2018}, imagery-based representations allow for direct inspection of relationships within a scene, bypassing the resource-intensive chaining required by propositional logic systems. This efficiency becomes critical as the complexity and volume of data increase, highlighting the potential of visual reasoning frameworks.

Despite these insights, current AI systems lack mechanisms for simulating internal mental imagery as part of their reasoning processes. While generative models can synthesize images from latent representations, they do not yet operate within a unified framework that links internal imagery to cognitive states or decision-making. Moreover, the source data for these generated images—i.e., what memory, knowledge, or perceptual cues they are based on—remains unspecified and disconnected from dynamic reasoning contexts.

In essence, the ability of the human brain to reconstruct experiences into mental images is key to flexible thinking and imagination. However, this mechanism is largely absent in state-of-the-art AI systems. Bridging this gap requires the development of new models that explicitly incorporate internally generated visual simulations into reasoning architectures, making mental imagery a first-class component of artificial cognition.

\section{Machine Thinking: The Proposed Framework}\label{sect4}

Our goal is to propose a framework for machine thinking, inspired by the way the human brain operates. Given the limited understanding of how the brain executes cognitive processes, we draw inspiration from recent advances in neuroscience.

\begin{figure}[ht!]
\centering
\includegraphics[width=5cm]{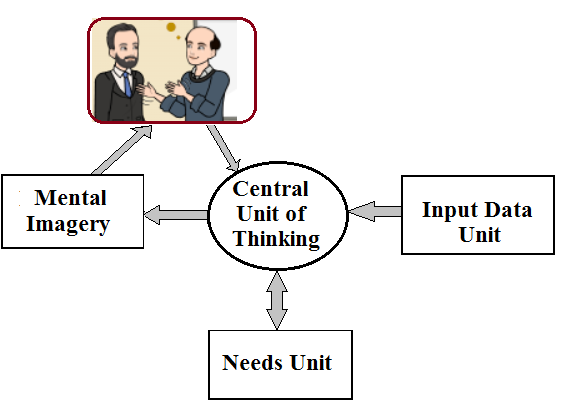}
\includegraphics[width=5 cm]{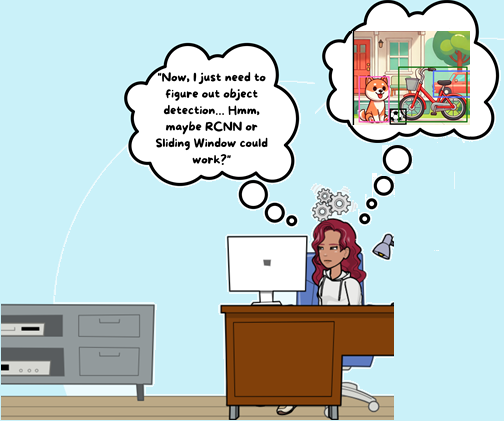}
\caption{(Left)Diagram of Machine Thinking algorithms. (Right) Output of Human Thinking.}
     \label{figure_r1}
\end{figure}

We define machine thinking as the process of inferring decisions and generating informative content from various data sources, with the results presented as natural language sentences and sketches~\cite{Huang2020}. These sketches represent a series of machine-generated mental images, mimicking a common aspect of human cognition (see Figure~\ref{figure_r1}).

The synoptic scheme of the proposed framework, illustrated in Figure ~\ref{figure_r1}, comprises a \textbf{Cognitive Unit of Thinking} which processes information received from the three auxiliary units: The \textbf{Needs Unit}, the \textbf{Input Data Unit} and the \textbf{Mental Imagery Unit}. The Cognitive Thinking Unit \textbf{(CTU)} synthesizes information from multiple sources and produce an output including both informative and decision-making content.

\subsection{The "Needs" Unit}

This unit contains data that may include:(1) A set of actions to be performed, typically scheduled as a result of prior reasoning. These actions are generated by the Cognitive Thinking Unit (CTU) and transferred to the Needs Unit for execution planning. (2) Knowledge acquired by the machine, representing internal goals or obligations.

For a human analogy, examples of knowledge include:\\
"At 8:00 AM, I must be at my desk."\\
"According to my agenda, I have a meeting at 10:00 AM with my PhD student in the next building."

Examples of actions to be performed, derived through reasoning, include:\\
"To get to the next building, take the stairs."\\
"Go to the Queens Building using the bike."\\

\subsection{The "Input data" Unit}

Is a unit which processes the acquired images, the heard audio signals and produces them as a set of natural-language sentences describing the received data. The recent advances in computer vision using neural approaches may be used such as YOLO for object detection and recognition ~\cite{Redmon2016} and DenseCap for image captioning ~\cite{Johnson2016}. In the same way, the audio signal is converted into useful sentences. This involves several key steps in the pipeline of Automatic Speech Recognition (ASR) using for example DeepSpeech and Wav2Vec 2.0 and Natural Language Understanding (NLU) using Transformers-based models (BERT, GPT).
The sens of touch is also integrated for input data. Indeed, by using the tactile sensors, machine learning, and natural language generation, the sense of touch can be effectively translated into meaningful sentences that describe the physical properties of objects and interactions.\\

\subsection{The "Mental Imagery" Unit}

The Mental Imagery Unit receives stimulus from the Cognitive Thinking Unit (CTU), draws imagined images and communicates the convenient ones to the (CTU) which refine the result of reasoning.\\

\subsection{The Cognitive Thinking Unit (CTU)}\label{subCUT}

Thinking which is the ability to reach logical conclusions on the basis of prior information is central to human cognition ~\cite{Goel2017}. Therefore, the advent of neuroimaging techniques has increased the number of studies related the study of examined brain function ~\cite{Bartley2019} and to the neural basis of deductive reasoning ~\cite{Wang2020}.

The Cognitive Unit of Thinking refers to a central processing module or system that is responsible for Interpreting sensory or internal stimuli, Initiating reasoning processes, Making inferences based on knowledge, Integrating information from memory, perception, and imagination and Triggering mental imagery when needed for simulation or problem-solving.

In this paper, in addition to inferring action from a set of data (sentences) and a specific need, we study how mental images may generate new information for the machine. 

\section{Generating Mental Images}\label{sect5}

After receiving a stimulus (high-level cognitive instruction) from the CTU (Cognitive Thinking Unit), the Mental Imagery Unit (MIU) generates and visualizes images, which are then perceived by the CTU. In the next step, reasoning is performed based on the content of these images.
The stimulus may be: \\
- A sentence or semantic description representing a situation (“Imagine a cup falling off a table”)\\
- Hypothetical query (“What if the door were locked?”)\\
- Goal-oriented stimulus ("make a mosquito net"). 

\subsection{Mental Image Generation: Timing and Content}

We assume that each piece of knowledge is stored in memory as a sentence and can be represented as a sketch derived from previously seen images. In computational terms, this implies the existence of a neural model with appropriate parameters capable of producing such visual representations.

When a new hypothesis (resulting from reasoning) is integrated into existing knowledge, the corresponding image is modified to incorporate the additional information. This enables the generation of continuously updated images as reasoning progresses and new knowledge is formed. The CTU then performs further reasoning based on these visualized images.\\

\begin{figure}[ht!]
\centering
\includegraphics[width=10cm]{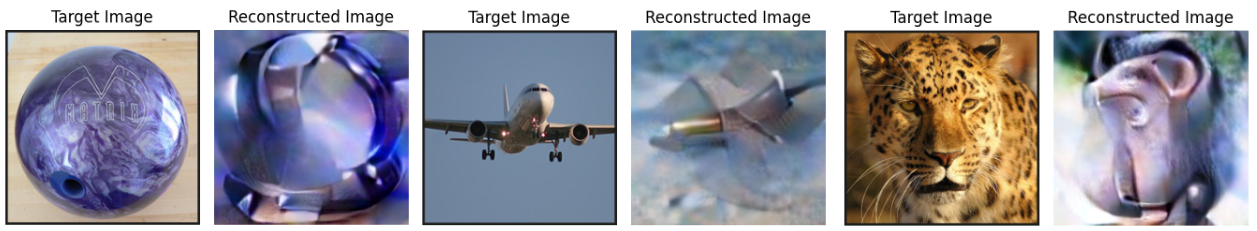}
\caption{Reconstruction of imagined images from brain activity ~\cite{Naoko 2024}.}
     \label{figure2_r}
\end{figure}

In ~\cite{Naoko 2024}, K. M. Naoko et al. proposed a deep neural-network for reconstruction of both seen images (i.e., those observed by the human eye) and imagined images from brain activity. Authors applied the reconstruction methods to the fMRI signals measured during imagery. They used the data measured while the subject imagined natural images and artificial shapes. Figure \ref{figure2_r} shows examples of such imagined images.

\begin{figure}[ht!]
\centering
\includegraphics[width=8cm]{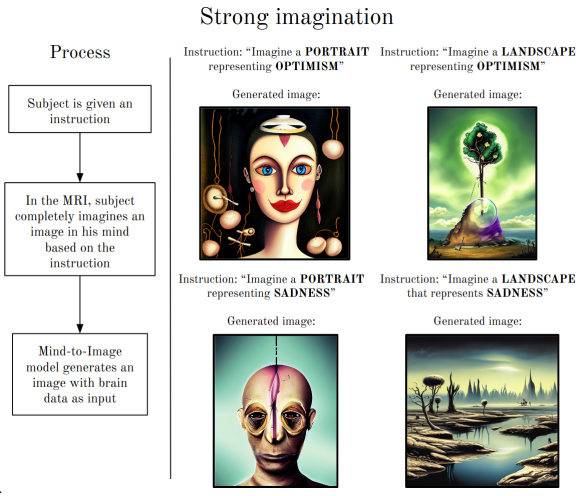}
\caption{Results from strong imagination. The subject is given an instruction on the fMRI screen, such as ”Imagine a portrait representing optimism”. Then, the subject purely imagines such image, giving an oral description. Then, the associated brain data is fed to the fMRI-to-Image model to produce a reconstruction ~\cite{Hugo 2024}.}
     \label{figure3_r}
\end{figure}

The ability to decode and reconstruct these imagined visuals offers a wide range of promising possibilities from aiding individuals with disabilities, to enhancing creative processes ~\cite{Hugo 2024}. In ~\cite{Hugo 2024}, authors train
a modified version of an fMRI-to-image model and demonstrate the feasibility of reconstructing images from two modes of imagination: from memory and from pure imagination (see figure \ref{figure3_r}).

\subsection{Thinking: A Computational Perspective}

In our mind, we can move in mental image representing a scene taking a certain time. This scene may change in the time and the brain is then exploring or simulating the communicated event. In \cite{Peebles2019}, D. Peebles tested for Mental scanning the classic study of Kosslyn et al. \cite{Kosslyn1978} in which people were required to memorise the locations of landmarks on a fictitious map. Participants were timed while carrying out the task and analysis of the response times (RTs) revealed a linear relationship between the distance travelled and the time taken
to reach the destination.  

\begin{figure}[ht!]
\centering
\includegraphics[width=8cm]{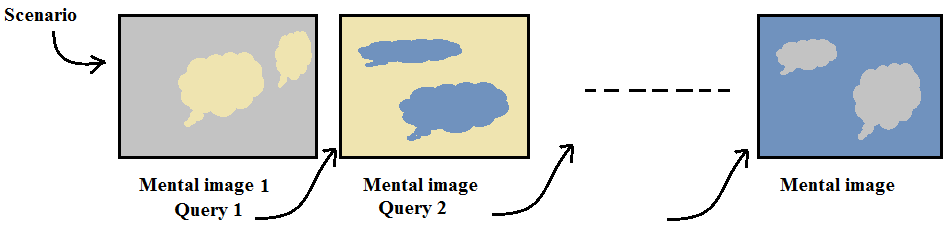}
\caption{Set of mental images generated from a scenario and reasoning queries associated.}
     \label{figure4_r}
\end{figure}

We consider thinking as the process of exploring possible ways to achieve an event by generating a time series of mental images. This process can be performed in different ways:\\
Once a sentence (representing an event) is communicated by the CTU unit, a set of mental images is generated and visualized, representing the progression from the initial state to the final state. These visualized images correspond to the unfolding of the event from beginning to end (see figure \ref{figure4_r}).
The generation and visualization of these images rely on contextual knowledge, inference, and the ability to introduce changes to the content of the images.
The change of the mental image content is made in order to add details, or to correct partially the event or change completely the content. 
Once the process of seeing and adjusting the content of mental images is complete, the event is scheduled as a set of elementary events corresponding to the set of generated mental images.

An emerging challenge is determining how a computational system can interpret and engage with internally generated mental images.
The understanding aspect is relatively straightforward, as the mental image is constructed conditionally based on a given linguistic input (i.e., a sentence describing an event or scenario).
To interact with the mental image, the system can perform reasoning operations by generating and evaluating queries such as: Is the depicted scenario physically or logically feasible?, What preconditions or resources are required to realize this configuration?, and other forms of inferential questioning. These interactions enable the system to assess plausibility, plan actions, and simulate outcomes within the mental imagery framework.

\section{Experiments}\label{sect6}

\subsection{Input Data: available and generated}

Input of the system are obtained using sensors. In our experiments, we consider only the camera which delivers images sequence of the behaviour scene. Images segmentation and captioning using Neural Models give a set of informative sentences which are the data of the Input Unit.

The Needs Unit contains the scheduled actions to be performed by the CTU unit.


For object Detection, we used Faster R-CNN, a deep learning model from the Torchvision library. Faster R-CNN analyzes the input image and detects various objects by outputting their bounding boxes, class scores, and other attributes. The model is loaded with pretrained weights trained on the COCO dataset, enabling it to detect a wide range of common objects. The detections are filtered based on a confidence threshold, ensuring only objects with high detection accuracy are considered.

\begin{figure}[ht!]
\centering
\includegraphics[width=5 cm]{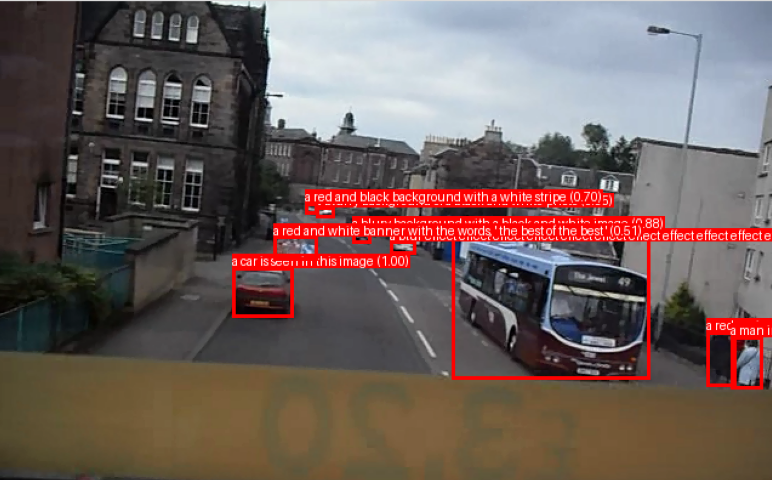}
\includegraphics[width=3 cm]{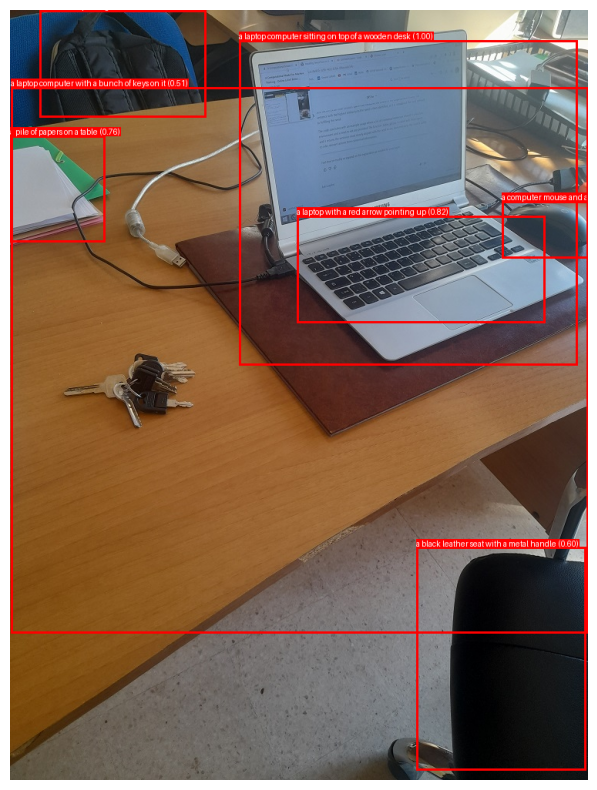}
\caption{(Left) Some of captions are: "a car is seen in this image", "a bus is driving down the street", "a red car is parked in a parking", "a man in a white shirt and black pants is seen through a red square". (Right) A sample of used image for captioning. Detected objects are illustrated with their associated bounding box.}
     \label{figure5_r}
\end{figure}

For each detected object, the bounding box coordinates are used to crop the corresponding region from the input image and the cropped region is passed to a BLIP (Bootstrapped Language Image Pretraining) model, a transformer based image captioning model, which generates a natural-language description of the object. This combination allows the pipeline to describe specific regions of an image, producing captions like "a red car is parked in a parking." as shown by figure ~\ref{figure5_r}(Left).

The following sentences are generated for the image shown by figure ~\ref{figure5_r}(Right).\\

sentences = [ a laptop computer sitting on top of a wooden desk, a computer mouse and a mouse pad, a box with a pair of gloves in it, a laptop with a red arrow pointing up, a pile of papers on a table, a black leather seat with a metal handle, a laptop computer with a bunch of keys on it.]

\subsection{Matching the Need and the Context Sentences: In the Cognitive Unit of Thinking}

Sentences representing the need, available in the Needs Unit, are communicated to the CTU unit. One possibility of thinking is to infer the response of the need based on available context based on the data of Input Unit. 

Given the context sentences defined as a list of sentences describing various objects (captions) in the image, the Need: A single sentence that expresses a requirement or goal. such as: ("I need the keys to open the door and go out"). The goal of the (CTU) unit is to find the sentence from the context that best matches or aligns with the need.

We propose for this task to use the "sentence-transformers" library, which is a Python framework for state-of-the-art sentence, text and image embeddings, to identify the most contextually relevant sentence from a list of sentences based on a specific "need" statement. 

In our implementation, we perform the following steps:\\
Load the pretrained model (all-MiniLM-L6-v2) from the SentenceTransformer library. This model converts text into high dimensional vector representations (embeddings). This model is chosen because of is  lightweight and optimized for tasks like sentence similarity, clustering, or semantic search.
Each context sentence is converted into a vector representation using the "model.encode()" method. The need sentence is also encoded into a vector. These embeddings are tensors, enabling efficient computation of similarities.

The cosine similarity between the need embedding and each context sentence embedding is calculated using (util.cos-sim()). 
The sentence with the highest similarity score is identified. The corresponding sentence is extracted as the best match for the given need.

As example, for the following context sentences, and having the 
need = "need the keys to open the door and go out", the similarity scores computed for all sentences are:\\
a bottle of water sitting on a table - Similarity: 0.0395\\
a desk with a chair and a laptop on it - Similarity: 0.1353\\
a table with a laptop and a bottle of water - Similarity: 0.1645\\
a white table with a red line on it - Similarity: 0.0807\\
a computer mouse and a mouse pad - Similarity: 0.1755\\
a box with a pair of gloves in it - Similarity: 0.1869\\
a laptop with a red arrow pointing up - Similarity: 0.1376\\
a pile of papers on a table - Similarity: 0.0662\\
a black leather seat with a metal handle - Similarity: 0.1675\\
\textbf{a laptop computer with a bunch of keys on it - Similarity: 0.4531}\\

A language model from Hugging Face's Transformers library is used in our case to infer appropriate actions based on given contextual observations and specific needs. The different steps performed are:

- Specify the used tokenizer and the model from Hugging Face's servers.\\
- Define the context: A set of sentences, each representing an observation in the environment.\\
- Define the Need: A set of sentences, each describes a specific requirement.\\
- Generate the actions based on the provided contexts and needs using the function $generate\_action\_falcon$.\\
- Combine all context sentences into a single string.\\
- Create a structured prompt which include: A description of the contextual observations, the specific need to be addressed and a request for an appropriate action.\\
- Use the combined prompt as the input to the model for generating a response. 

The combined prompt is tokenized into numerical representations using the tokenizer. The tokenized input ensures compatibility with the model and facilitates efficient processing. The model generation method is called to produce a response based on the prompt. The parameters for generation include the max length of the generated text to prevent overly verbose outputs and the randomness in the response (lower values make it more deterministic). The generated text is decoded back into a human-readable format.

\begin{figure}[ht!]
\centering
\includegraphics[width=2.0 cm]{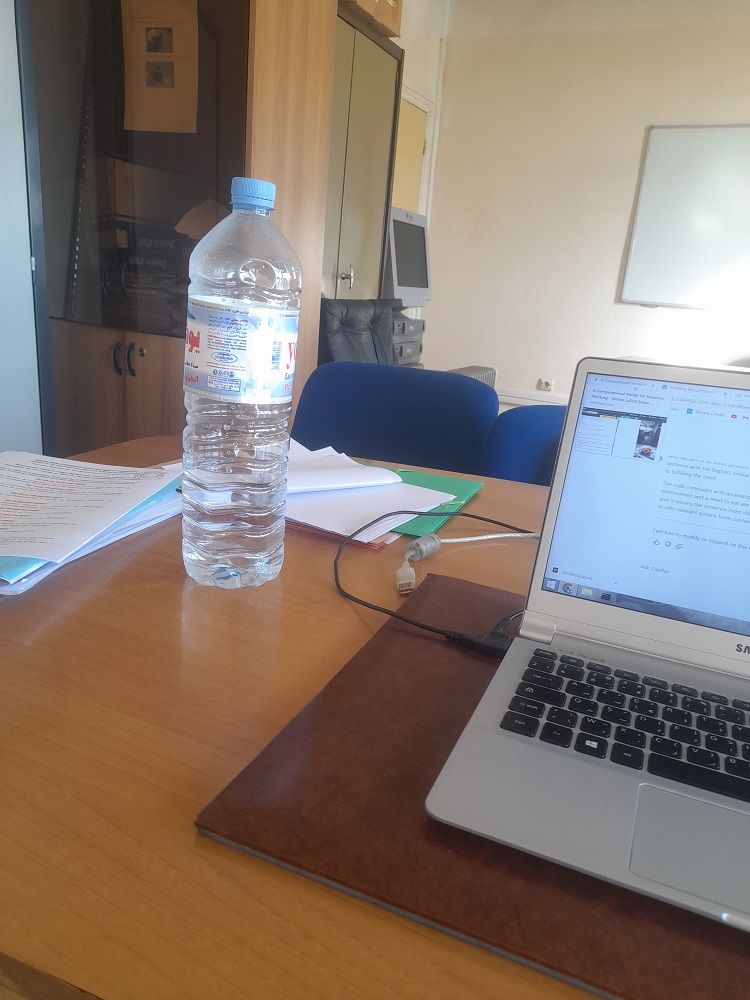}
\includegraphics[width=2.0 cm]{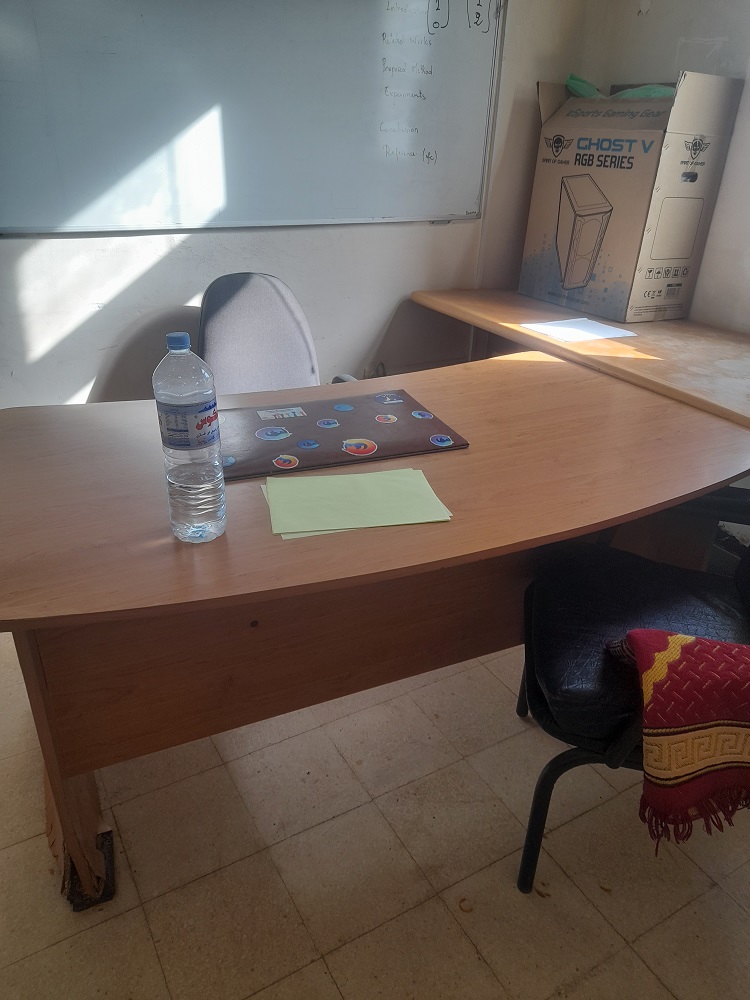}
\includegraphics[width=2.0 cm]{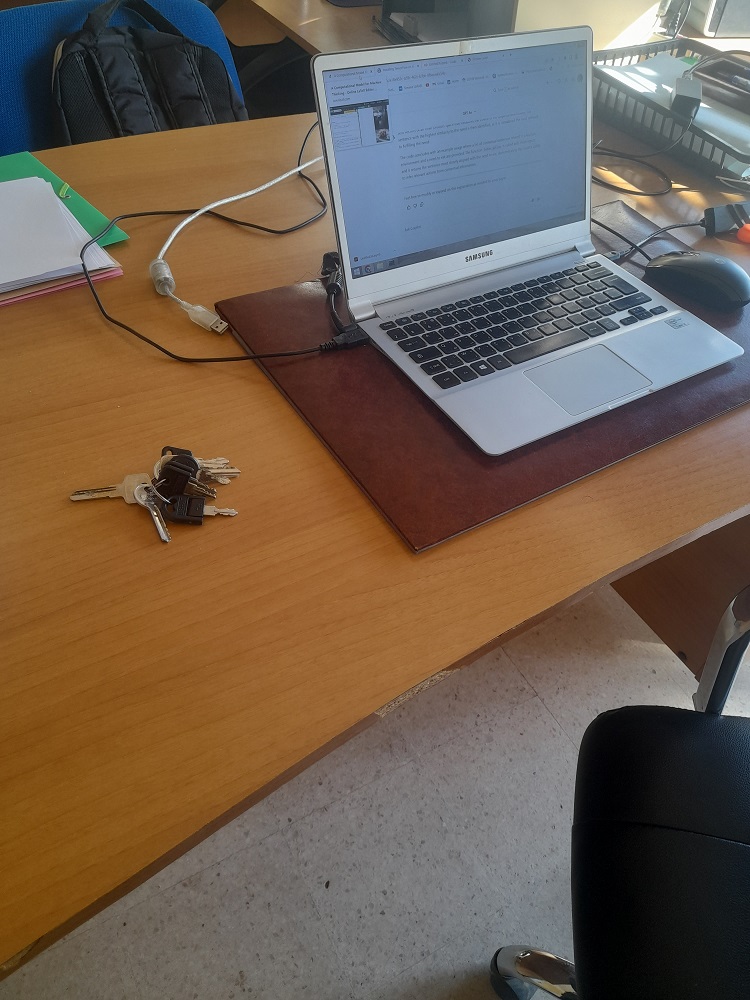}
\caption{A sample of used image sequence for captioning.}
     \label{figure6_r}
\end{figure}

The following captions (contexts) are given respectively for images of figure ~\ref{figure6_r}.\\

contexts = [\\
    "a laptop computer with a bunch of keys on it",
    "a bottle of water sitting on a table",
    "a desk with a chair and a laptop on it",
    "a pile of papers on a table"
]\\
We assumed that Need = ["need the keys to open the door and go out"].

As a result, we obtained the generated Action: "Pick up the keys and open the door - Take the keys and unlock the door - Use the keys to open the door and go out".

For a second Need : ["I need to drink water to stay hydrated"], the generated action is "Take a sip of water from the bottle on the table".
  
\subsection{Draw the Mental Images}

This experiment investigates the integration of a Cognitive Thinking Unit (CTU) and a Mental Imagery Unit to simulate the process of human-like mental visualization based on action sequences. The goal is to assess the system's ability to generate meaningful, sketch-style mental images from natural language descriptions of sequential actions. This mimics how humans imagine future scenarios during reasoning or planning.

The system is composed of two main components:

1. Cognitive Thinking Unit (CTU): This unit simulates the decision-making process by generating a predefined sequence of natural language action descriptions. The selected action plan in this experiment included the steps: "a man takes the keys which are on the desk", "the man goes towards the door", "the man opens the door".

2. Mental Imagery Unit: This module uses a pre-trained Stable Diffusion model (`runwayml/stable-diffusion-$v1-5$`) to generate visual representations of the actions. The generated images are then processed through a sketch transformation pipeline that converts photorealistic outputs into grayscale sketches using a pencil-sketch filter based on Gaussian blurring and image division techniques.

For each action, one mental image was generated and converted into a sketch, mimicking a stylized internal visualization process similar to human imagination.

The system successfully produced a sequence of mental sketches corresponding to each action in the plan. The generated sketches captured the essential spatial and semantic elements of each action: Object interaction, Spatial transition, Interaction with environment and Scene transition.

\begin{figure}[ht!]
\centering
\includegraphics[height=2.0 cm]{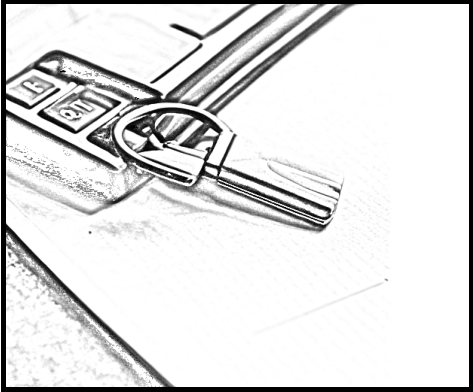}
\includegraphics[height=2.0 cm]{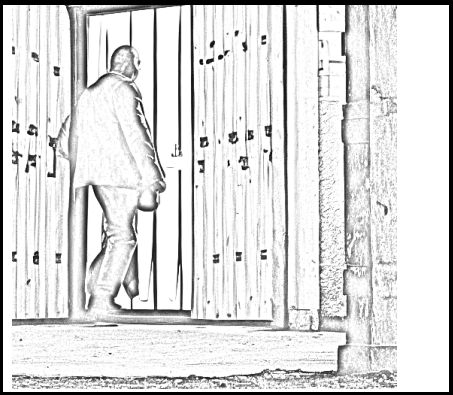}
\includegraphics[height=2.0 cm]{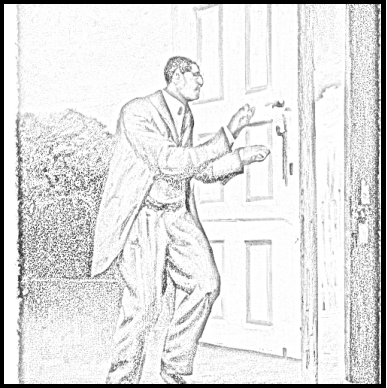}
\caption{Mental images generated from the three prompts.}
     \label{figure7_r}
\end{figure}

These results (see figure \ref{figure7_r}) show that the integration of text-to-image diffusion models with sketch-based abstraction effectively emulates a simplified form of mental simulation, where visual reasoning can be initiated from natural language actions. The sketch style also supports cognitive tasks by highlighting structural content without unnecessary visual noise.

This experiment demonstrates that mental imagery generation from action sequences is a viable computational approach to support machine reasoning. By abstracting scenes into sketches, the system aligns more closely with how humans imagine events in simplified visual terms. While the current system uses predefined actions, future extensions could incorporate dynamic action planning and goal-oriented scene evolution.

\begin{figure}[ht!]
\centering
\includegraphics[height=2.0 cm]{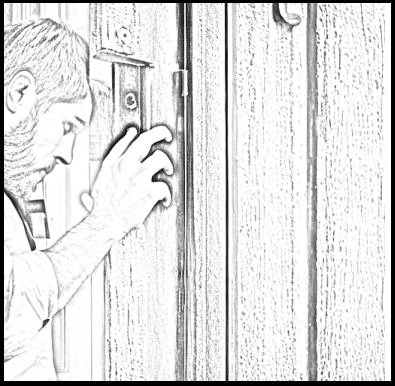}
\includegraphics[height=2.0 cm]{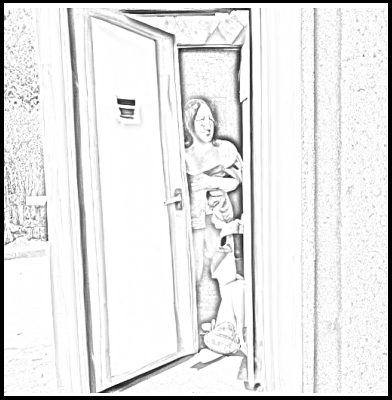}
\includegraphics[height=2.0 cm]{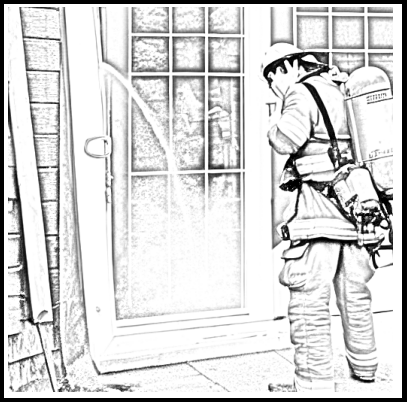}
\caption{The new mental images generated receiving from CTU unit new prompts.}
     \label{figure8_r}
\end{figure}

Reasoning can be applied to each image by asking logical questions—for example, 'What if the key doesn't open the door?' The CTU unit proposes a new action:"the person is nervous" , then a new mental image is then generated. The same process is repeated and a new possible actions are generated: "the person breaks down the door", "the person call the firefighters for they open the door". Mental images generated are shown by figure \ref{figure8_r}.

\section{Conclusion}\label{sect7}

In this paper, we addressed the ambitious challenge of enabling machine thinking by proposing a comprehensive framework inspired by recent advancements in neuroscience, deep learning, and artificial intelligence. While significant progress has been made in developing systems capable of processing sensory data, reasoning, and interacting with humans, our approach goes beyond existing models by integrating the concept of a Mental Image to simulate machine cognition.  \\

We presented the limitations of current models, including their reliance on explicit queries, domain specificity, and the absence of intuitive reasoning mechanisms are discussed. These gaps motivated the design of our framework which consist of a Cognitive Unit of Thinking (CTU) and three auxiliary units: The Needs, Input Data, and Mental Image which work in synergy to mimic human cognitive processes. This novel architecture enables machines to reason, plan, and infer decisions autonomously, utilizing both sensory inputs and internally generated representations.  \\

The proposed framework have been implemented and the results demonstrated the framework's potential to bridge perception, reasoning, and imagination, thereby addressing key limitations of existing systems.

\end{document}